\documentclass[graybox]{svmult}

\usepackage{type1cm}        % activate if the above 3 fonts are
\usepackage{makeidx}         % allows index generation
\usepackage{graphicx}        % standard LaTeX graphics tool
\usepackage{multicol}        % used for the two-column index
\usepackage[bottom]{footmisc}% places footnotes at page bottom

\usepackage{newtxtext}       % 
\usepackage{newtxmath,comment}       % selects Times Roman as basic font
\usepackage{xspace}
\usepackage{microtype}
\usepackage{subfigure}
\usepackage{color}
\usepackage{url}
\usepackage{lipsum}
\usepackage{fancyhdr}
\usepackage{setspace}
\usepackage[table]{xcolor}
\usepackage{amsfonts,amsmath}
\usepackage{bm}
\usepackage{enumitem,centernot}
\usepackage{dsfont}
\usepackage[ruled,vlined]{algorithm2e}
\usepackage{algpseudocode}
\usepackage{color, colortbl}
\usepackage{tabularx}
\usepackage{multirow}
\usepackage{pifont}
\usepackage[labelformat=empty]{subcaption}
\usepackage{caption}
\usepackage{float}
\usepackage{diagbox}
\usepackage{bbm}
\usepackage[numbers]{natbib}

\makeindex             % used for the subject index
\begin{document}

\title*{The Game-Theoretic Symbiosis of Trust and AI in
Networked Systems}
\author{Yunfei Ge and Quanyan Zhu}
\institute{ Yunfei Ge \at New York University, \email{yg2047@nyu.edu} \and Quanyan Zhu \at New York University \email{qz494@nyu.edu}
}
\maketitle

\abstract{This chapter explores the symbiotic relationship between Artificial Intelligence (AI) and trust in networked systems, focusing on how these two elements reinforce each other in strategic cybersecurity contexts. AI's capabilities in data processing, learning, and real-time response offer unprecedented support for managing trust in dynamic, complex networks. However, the successful integration of AI also hinges on the trustworthiness of AI systems themselves. Using a game-theoretic framework, this chapter presents approaches to trust evaluation, the strategic role of AI in cybersecurity, and governance frameworks that ensure responsible AI deployment. We investigate how trust, when dynamically managed through AI, can form a resilient security ecosystem. By examining trust as both an AI output and an AI requirement, this chapter sets the foundation for a positive feedback loop where AI enhances network security and the trust placed in AI systems fosters their adoption.}

% \input{introduction/introduction}

% \nocite{*} % remove this line to add only cited references

\section{Trust in Networked Systems}
The rapid development of network systems has been a catalyst for innovations such as 5G communications, edge computing, and network slicing \cite{foukas2017network}, driving the transformation of Industry $4.0$ \cite{ghobakhloo2020industry} and introducing new services for critical infrastructures. This has led to a more interconnected and expansive network environment, where Information Technology (IT) and Operational Technology (OT) networks converge, creating large, hybrid systems with heterogeneous devices \cite{murray2017convergence}. However, this evolution has also expanded the attack surface, with threats becoming more sophisticated and stealthy. Techniques such as Advanced Persistent Threats (APTs) pose significant challenges, making the security of these complex, connected systems increasingly difficult to manage. Despite the transformative benefits of these advancements, ensuring the security and resilience of networked systems remains a critical concern.

\begin{figure}[!t]
    \centering
    \includegraphics[width=0.9\linewidth]{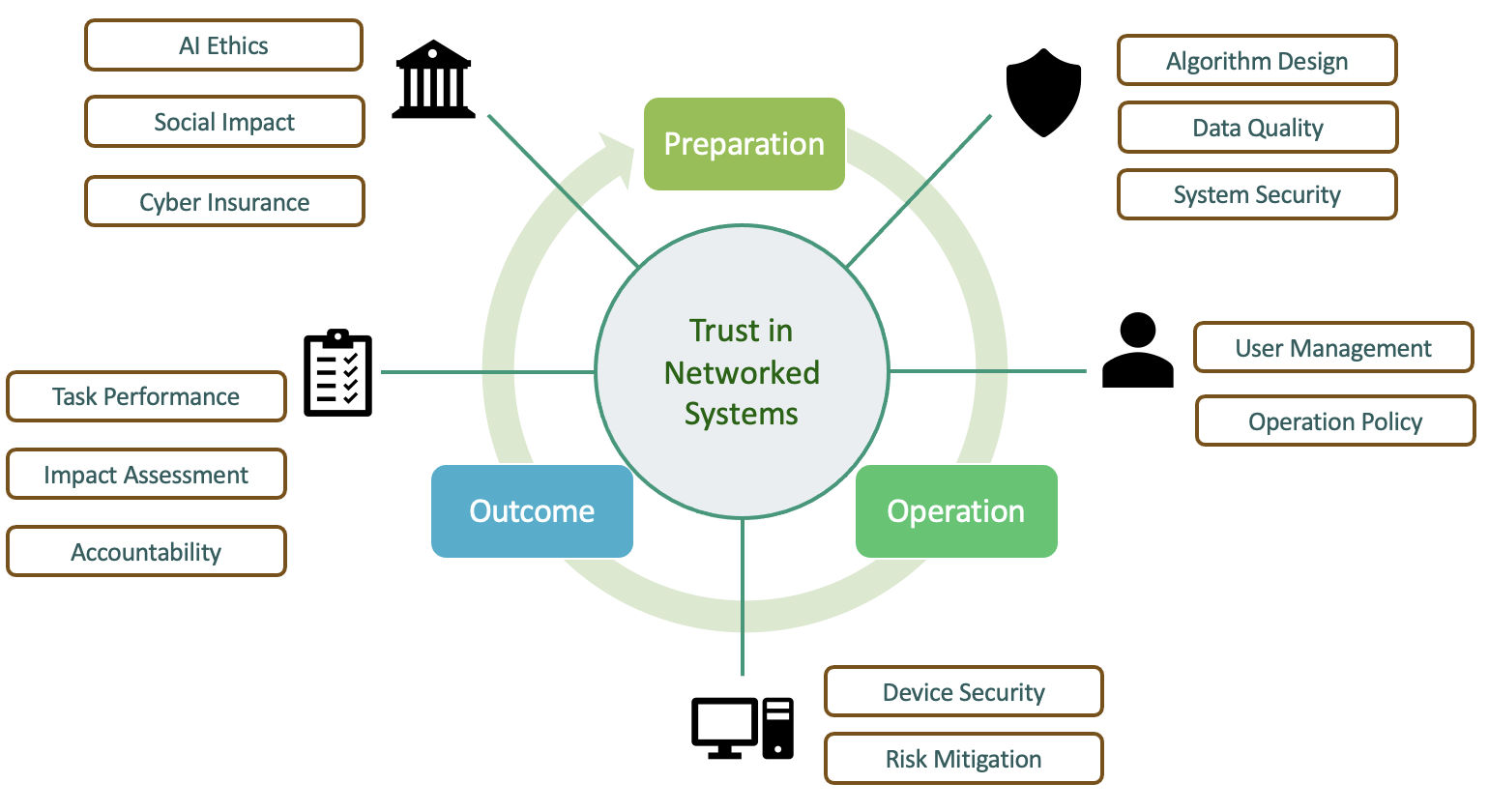}
    \caption{Trust is integral to all stages of a networked system. A trustworthy network system ensures reliability from preparation and operation through to the outcomes.}
    \label{fig:trustsec}
\end{figure}

At the core of network security is trust. When trust is compromised, it can lead to devastating consequences, including security breaches, data loss, and a loss of confidence in the integrity of systems and services \cite{ge2023zero}. Trust permeates every phase of a network’s lifecycle—from its initial setup to its operational outcomes, as illustrated in Figure~\ref{fig:trustsec}. It manifests across multiple dimensions \cite{cho2015survey}. Firstly, trust in network policy is paramount, particularly when it comes to penetration testing and vulnerability assessments \cite{ge2024mega}. Poorly constructed or untrustworthy policies can result in security gaps or malfunctions in network operations. Secondly, trust in identity is crucial, as it underpins access control mechanisms \cite{ge2022mufaza, ge2023gazeta}. Without confidence in the identity of entities within the network, malicious actors could easily infiltrate, making it impossible to ensure that only legitimate users or devices are granted access. Lastly, trust in system performance is vital, especially in critical infrastructures where reliability and accountability are non-negotiable \cite{ge2022accountability}. If a system's performance falters, users may lose trust in its reliability, discouraging adoption and threatening its long-term viability. Thus, addressing and comprehensively understanding the dimensions of trust in modern networked systems is vital for safeguarding them.

Existing approaches to establishing trust in networked systems are often inadequate. Trust in network policy, for example, frequently relies on perimeter-based security models, which are insufficient for addressing insider threats \cite{stafford2020zero}. Trust in identity typically hinges on rule-based checks and encryption techniques, leaving systems vulnerable to identity fraud, credential theft, and other manipulations \cite{yang2019faster,matheu2020survey}. Moreover, trust in system performance is under constant threat from increasingly sophisticated attacks, such as APTs \cite{huang2020dynamic}, which can manipulate system behavior undetected, as exemplified by attacks like Stuxnet \cite{kushner2013real}. These limitations point to the urgent need to rethink how trust is utilized, measured, and protected in networked environments. The central research question thus becomes: How can we redefine trust in a way that better accounts for the dynamic, strategic nature of interactions in modern network systems to improve security?

\subsection{Deception and Trust}
Trust and deception are inherently intertwined. Cyber deception, by design, seeks to manipulate trust—either by instilling a false sense of trust where there should be none or by fostering distrust where trust is warranted \cite{pawlick2021game}. The goal of deception is often to create a misleading narrative that either encourages the deceivee to place trust in a compromised entity or distracts them from recognizing legitimate threats \cite{zhang2018hypothesis,zhang2020game}. Defensive cyber deception, for instance, frequently employs techniques like honeypots \cite{huang2019adaptive}, designed to lure attackers into engaging with decoy systems that mimic production environments. This misleads the attacker into believing that they have infiltrated a valuable target when, in reality, they are interacting with a controlled system. Conversely, attackers may deploy deceptive tactics, such as generating false alerts to overwhelm system operators, thereby obscuring legitimate threats within a sea of noise. This exploits human cognitive vulnerabilities, particularly their limited attention, to erode the operator’s ability to effectively trust the validity of alerts \cite{kim2024human}.

Trust lies at the heart of deception. To fully understand deception in networked systems, it is essential to examine trust not just as a static or random variable but as a strategic interaction that can be shaped, manipulated, and exploited \cite{ge2024symbiosis}. Current models of trust in computing systems often treat it as an exogenous factor—something that exists outside the system, randomly determined, and subject to estimation \cite{sharma2020towards}. However, this view fails to capture the strategic nature of trust interactions in environments where adversarial behaviors and deceptions are prevalent. Trust can be manipulated, forged, or undermined, and thus requires a more nuanced approach that considers its strategic dimensions \cite{ge2023zero}.

\subsection{A Strategic Approach to Trust in Networked Systems}

We propose reframing trust as a dynamic and controllable interaction, rather than a static variable. This shift in perspective is critical for addressing challenges in zero-trust architectures, cyber deception, misinformation campaigns, and evasion tactics used by attackers. Trust is not merely something to be estimated or passively observed—it is an interactive process that can be deliberately shaped by the entities that control it. By understanding trust as a strategic component, we can begin to develop models that account for adversarial behaviors and the manipulation of trust. This includes both recognizing when trust is being falsified and understanding how to foster rightful trust in networked environments. 

To advance this understanding, we first examine traditional approaches to modeling trust in computing systems. Then, we introduce the  of trust as a strategic interaction—one that requires models that endogenize the behaviors of both trusted and untrusted entities, as well as those who seek to manipulate or exploit trust. This strategic approach diverges from existing models by recognizing that trust is  exogenous to the system and that it is not just a variable shaped by the entities outside the system. Understanding the endogenized strategic framework is essential for developing more robust cybersecurity strategies that can adapt to the increasingly complex and deceptive nature of modern networked environments.

The strategic understanding of trust is directly applicable to several critical areas, including zero-trust systems,  cyber deception, combating misinformation, and attack evasion. In each of these contexts, trust is not merely a static attribute but a dynamic element that can be manipulated, forged, or strategically withheld, depending on the entities controlling it. Let’s explore how this strategic approach to trust applies to each of these areas:

\subsubsection{Zero-Trust Systems}
In zero-trust architectures, the traditional model of implicit trust based on network boundaries is abandoned. Instead, every user, device, and application is continuously authenticated and validated, regardless of their location within or outside the network perimeter \cite{stafford2020zero}. The strategic nature of trust plays a key role here, as trust is not automatically granted but must be earned through continuous verification.

When trust is viewed as strategic and controllable, zero-trust systems can be designed to dynamically adjust trust levels based on contextual information. For instance, behavioral analytics, adaptive authentication, and risk-based access control can be employed to assess the trustworthiness of users or devices in real-time \cite{ge2023gazeta}. The entity controlling trust in this scenario—the system administrator or security mechanism—must constantly evaluate whether trust should be granted, reduced, or revoked based on the observed behavior of the network's participants \cite{ge2023scenario}. This strategic adjustment of trust helps to prevent both internal and external threats, ensuring that no entity within the system is blindly trusted.

\subsubsection{Cyber Deception}
In cyber deception, attackers attempt to manipulate trust to gain unauthorized access or influence the target's perception of the system. For example, an attacker may use phishing techniques to forge trust by impersonating a legitimate user or entity \cite{huang2023advert}. Similarly, defensive cyber deception uses techniques like honeypots \cite{li2024symbiotic} to mislead attackers into trusting decoy systems.

A strategic understanding of trust is crucial in cyber deception, as it allows for the design of systems that can both detect and exploit the adversary’s manipulation of trust. In a defensive context, security teams can manipulate the trust perceptions of attackers by creating deceptive environments that appear legitimate. For attackers, the challenge is in controlling trust in ways that can evade detection while gaining access to critical resources. Understanding trust as a manipulable element on both sides of the deception equation provides deeper insights into how to defend against or employ deception more effectively.

\subsubsection{Combating Misinformation}
Misinformation campaigns thrive on the manipulation of trust. False information is often designed to appear trustworthy, aiming to mislead audiences or undermine confidence in reliable sources \cite{yang2023designing}. Whether it’s disinformation spread through social media, fake news websites, or deepfakes, the central tactic involves forging a false sense of trust in misleading content.

When trust is viewed strategically, combating misinformation involves identifying and disrupting the mechanisms by which false trust is established. This includes using algorithms to verify sources, flagging inconsistencies, and providing contextual information to restore rightful trust in legitimate sources. Strategic trust models can also help to identify patterns of misinformation dissemination and predict how trust in certain types of content is manipulated. Moreover, defensive strategies can be devised to reinforce trust in accurate information while delegitimizing untrustworthy content.

\subsubsection{Attack Evasion}
In attack evasion, attackers attempt to manipulate the trustworthiness of their activities to bypass security mechanisms, such as intrusion detection systems (IDS) or antivirus software. For example, attackers may use techniques such as polymorphic malware, which alters its appearance to evade detection, or low-and-slow attacks that operate under the radar of traditional security tools.

From a strategic trust perspective, security systems must continuously adapt their trust evaluations to anticipate and respond to such evasive techniques. Trust in system behavior must be continuously reassessed based on the detection of anomalies, unusual patterns, or contextual data. By strategically controlling trust, systems can be designed to detect subtle deviations in normal behavior that might indicate an attack. For attackers, the challenge is to manipulate trust in such a way that their actions remain undetected while avoiding suspicion.

\section{Symbiotic Relationship Between AI and Trust}
\label{sec:sym}
Artificial Intelligence (AI) is changing how we analyze trust in networked systems by tackling the complexity, scale, and constant evolution of modern infrastructures. AI can process large volumes of log files and security alerts, improving both the speed and accuracy of trust assessments across networks. With advancements in processing power, AI now handles vast amounts of historical and real-time data, making it possible to assess the trustworthiness of users and devices more precisely.

AI systems also integrate expert knowledge from established security frameworks, such as OWASP, MITRE ATT\&CK, and the CVE database, which provides them with reliable insights into network vulnerabilities. This combination of expert knowledge and real-time data allows AI to make trust evaluations that are both robust and nuanced, surpassing traditional methods that rely on static, manual updates.

AI's adaptability and automation are crucial in a field where network configurations, users, and topologies change constantly. By monitoring networks around the clock and responding rapidly to threats, AI supports continuous trust management. Effective trust assessment goes beyond evaluating current actions to predict future behavior and intentions. For example, game-theoretic models \cite{zhu2018game, pawlick2019game, li2022role} allow AI to anticipate possible interactions between users, attackers, and defenders within a network, making trust management adaptive and forward-looking rather than solely reactive.

Machine learning methods like reinforcement learning and meta-learning further strengthen AI’s ability to adapt trust mechanisms in cybersecurity \cite{ge2023scenario}. These approaches help AI analyze past incidents to automatically adjust security policies, moving away from manual, reactive processes typical in Security Operations Centers (SOC). Reinforcement learning, in particular, allows AI to evolve trust policies in real-time, creating a self-sustaining security framework where systems continuously refine their defenses without waiting for human input \cite{apruzzese2023role}.

Despite its benefits, AI in trust analysis also presents challenges, particularly around transparency and ethics. Many AI models operate as ``black boxes'', meaning their decision-making processes aren’t easily understood, which can be risky when trust decisions impact critical infrastructure, such as power plants. Additionally, AI-driven decisions can inadvertently carry biases from training data, raising fairness and accountability concerns.

Trust in AI systems is essential for their adoption in security applications. Strong governance frameworks that define clear standards for AI development and deployment can help build this trust \cite{ge2024attributing}. These frameworks should be flexible and evolve alongside technological advancements. Tools like liability measures and insurance can address errors or unintended consequences, while transparent accountability mechanisms help ensure that ethical concerns are responsibly managed.

The relationship between AI and trust is mutually reinforcing. AI improves trust analysis in network security by driving strategic security enhancements and autonomous resilience, while the trustworthiness of AI affects its safe and effective application. This relationship can be seen as a ``meta-game'', where the capabilities of AI and the trust placed in it influence each other. In a positive equilibrium, AI strengthens trust in network security, and high trust in AI accelerates its integration. Conversely, low trust can limit AI’s role in security applications, emphasizing the need for strong governance and ethical oversight as AI becomes central to security and trust analysis.

In this chapter, we explore the synergy between AI and trust, aiming to achieve a balanced point where they positively reinforce each other. By focusing on AI and trust in networked cybersecurity, we investigate their interdependence and propose methods such as game theory and learning theories to provide new insights into trust in networked systems. Additionally, we propose governance frameworks that enhance the responsible use of AI systems, providing a foundation for strategic network security, autonomous resilience, and responsible AI governance.

The chapter is organized as follows. Section \ref{sec:sym} explores the symbiotic relationship between AI and trust, examining how AI transforms trust management through data processing, integration with established frameworks, and real-time adaptability. Section \ref{sec:game} discusses the use of game theory and AI for trust modeling and evaluation, reviewing techniques that focus on metrics, target entities, and evaluation methods. It covers policy-based and reputation-based approaches, as well as game-theoretic frameworks, to demonstrate how trust can be modeled as a strategic game. This approach enables systems to respond dynamically to adversarial behaviors, using Bayesian updates and strategic incentives to support resilience in zero-trust environments. Section \ref{sec:strengthen} addresses the vulnerabilities of AI algorithms and the role of game theory in enhancing AI trustworthiness. It includes a case study on an AI-driven traffic management system, illustrating how game-theoretic principles and red teaming strengthen trust and resilience in critical infrastructures. Section \ref{sec:conclusion} concludes the chapter, summarizing key insights and future directions.

\begin{figure}[!t]
    \centering
    \includegraphics[width=0.65\linewidth]{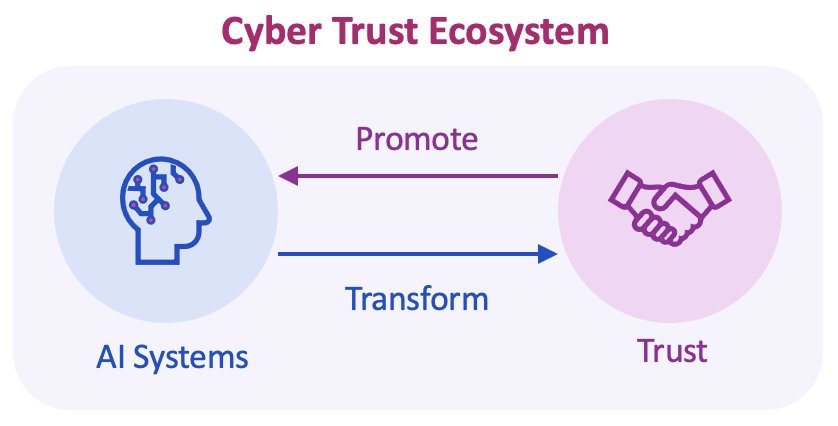}
    \caption{The symbiotic relationship between AI and Trust forms a cyber trust ecosystem, with each reinforcing the other.}
    \label{fig:eco}
\end{figure}

\section{Role of Game  Theory in Trust and AI}
\label{sec:game}
Game theory serves as the crucial link that bridges the gap between trust and AI, offering a structured approach to integrate these two domains. First, the strategic trust framework is inherently compatible with game-theoretic models. In a network environment, trust evaluation often involves adversarial agents who strategically manipulate the system to achieve their goals. Game theory allows us to model these agents’ behaviors, motivations, and interactions, enabling a deeper understanding of how trust can be established or undermined \cite{zhu2024foundations,manshaei2013game,zhu2018game}. By incorporating incentives, strategies, and objectives into the trust evaluation process, game theory provides a more targeted approach. Instead of evaluating trust by examining every possible scenario, which can be overwhelming, this framework directs attention toward the most relevant strategic interactions, making trust management more efficient and focused.

Second, game theory is not just a theoretical and analytical tool in economics but also an integral part of AI, and AI systems themselves can be leveraged extensively in trust management \cite{kamhoua2021game}. AI enables the transformation of raw data, past experiences, and domain-specific knowledge into actionable models for evaluating trust. With AI, decisions about trust management can become more data-driven and context-aware. AI algorithms can adapt to new information and identify patterns that humans might miss, making them invaluable for dynamically managing trust in complex and dynamic networks.

However, AI systems are not infallible. They are prone to errors, particularly when faced with unexpected inputs, adversarial attacks, or shifts in data distributions. This vulnerability makes it critical to develop robust AI methods that can defend against uncertainties and adversarial manipulation. Game-theoretic approaches have been used to address this challenge by framing the problem as a zero-sum game between the AI system and potential adversaries. In this context, the AI must maximize its robustness against data distribution shifts and adversarial inputs, while the adversary seeks to exploit weaknesses. These game-theoretic methods help in building more resilient AI systems that not only perform well in trusted environments but can also withstand attempts to undermine them.

At a higher level, as illustrated in Figure 2, there is a symbiotic relationship between AI and trust. AI technologies are transforming the way we evaluate and manage trust, and at the same time, trust is essential for the wider adoption and acceptance of AI as a reliable tool. This dynamic interaction can be modeled as a best-response scenario, where advancements in AI prompt improvements in trust management, and vice versa. When the two fields evolve in isolation, we risk falling into an equilibrium where distrust in AI limits its potential, stifling its role in transforming trust management systems. This is particularly dangerous in critical domains like network security or autonomous systems, where a lack of trust could delay the adoption of powerful AI-driven solutions.

To avoid this unfavorable equilibrium, it is essential to create an ecosystem where trust and AI mutually reinforce each other. This involves establishing governance mechanisms that promote the responsible use of AI in trust management, ensuring that AI tools are transparent, accountable, and reliable. When trust in AI is elevated, it, in turn, enhances AI’s ability to transform trust management, creating a virtuous cycle. The design of such an ecosystem can be informed by game theory, which offers a framework for understanding and optimizing strategic interactions. By using game-theoretic insights, we can craft policies that drive a positive equilibrium—one in which AI and trust grow hand-in-hand, reinforcing each other’s strengths, and leading to more secure and resilient systems.
 
This chapter focuses on the game theory-centric role in trust assessment and the trustworthiness of AI. Game theory aids in formalizing the adversarial interactions AI systems face from potential attackers while managing the trust mechanisms used by defenders. We use AI-driven traffic management as a case study, where game theory models the dynamics between the adversaries (who seek to disrupt traffic flow through data poisoning and model evasion) and defenders (who implement adversarial training and anomaly detection to maintain system integrity). By strategically assessing trust at every interaction, game theory enables us to optimize how AI systems manage vulnerabilities and how they respond to dynamic, adversarial threats.

In this section, we review the trust model and  methods. In the first part, we examine trust modeling, focusing on the key attributes that define trust in networked systems, such as the target of trust, metrics used for evaluation, and how trust information is collected and assessed. We discuss how trust-based decision-making is critical in environments where adversaries may deceive or mislead, especially in cybersecurity scenarios where entities may try to gain unauthorized access by manipulating trust metrics.

In the second part, we analyze trust management methods, exploring both policy-based and reputation-based approaches. We describe how policies can be defined to enforce trust through credentials, attribute checks, and incentive mechanisms, ensuring that only trusted entities can access system resources. Similarly, reputation-based methods use historical behaviors and third-party recommendations to establish trustworthiness, but these approaches are vulnerable to false data, making them less reliable in dynamic environments.

At the last subsection of this review, we introduce game-theoretic trust evaluation. This approach models trust as a strategic game between adversaries and defenders, where both parties adjust their strategies dynamically. Game theory allows us to model trust relationships in more complex, adaptive environments, enabling systems to anticipate adversarial behaviors and optimize trust management processes based on the evolving interaction between trusted and untrusted entities. This section will also explore how Bayesian updates and strategic incentives contribute to a more resilient trust framework, especially in zero-trust environments.

\subsection{Trust Modeling and Management}

Trust plays a pivotal role in networked system security. It has been extensively studied in various fields such as psychology, economics, political science, sociology, and computer science \cite{cho2015survey}. Essentially, trust refers to the degree of confidence an entity has in the expected behavior of another entity \cite{yan2008trust}. The trust-based decision-making is significant when there is a possibility of deception by the adversary, as in the case of cyber security. In such scenarios, attackers may intentionally mislead or conceal information to gain strategic advantage. Therefore, it is essential to understand the definition, metrics, and evaluation techniques of trust to create an efficient framework.  Table~\ref{tab:trust} illustrates the necessary attributes to design a trust-based framework in networked systems.

\begin{table}[t!]
\centering
\caption{Trust definition attributes in computer networks.\label{tab:trust}}{%
\begin{tabular}{@{}|c|c|@{}}
\hline
  \rowcolor{gray!80}
\textbf{Attributes}     & \textbf{Explanation}   \\
\hline
Target  & Who is the entity that will be evaluated\\\hline
Metric & What is the metric that is used to measure trust\\\hline
Collection & What information is collected to calculate trust\\\hline
Evaluation & How to evaluate trust\\\hline
Purpose  & How trust will be used in decision-making\\\hline
Management & How to manage the trust information in the system\\
\hline
\end{tabular}}{}
\end{table}

\subsubsection{Target of Trust} 
Different from transitional perimeter-based security, we expands the target of trust to every component in the network. Trust decisions will be based on not only the trustworthiness of the requiter but also on the device and environment where the data flow takes place. The granularity of the target  depends on the computational capability and the need of the system. 

\subsubsection{Metric of Trust} 
Trust-based decisions adopt a metric to measure the trustworthiness of the entity and provide risk analysis for policy decisions. In this chapter, we refer to this metric as the trust score (TS). To be specific, we formalize the trust score of an entity at the current time as the probability that the entity is non-adversarial to the system. Let $\theta\in\Theta$ be the attributes of the entity $i$, and denote the non-adversarial attributes set as $\Theta_T$. Formally,
\begin{definition}[\textbf{Trust Score}]
The Trust Score (TS) of the entity $i$ at time $t$ is defined as the probability that the entity is non-adversarial to the system:
\begin{align}
    TS^t(i):= \Pr(\theta^t_i\in\Theta_T) \in [0,1],
    \label{eq:ts}
\end{align}
where $\theta_i^t$ is the attributes of entity $i$ at time $t$. 
\label{def:TS}
\end{definition}
It should be noted that in practice, trust is multi-faceted and the attribute $\theta$ can be a multi-dimensional vector where each entry represents different trust attributes.

\subsubsection{Collection and Evaluation of Trust} 

We adopt the categories from Bonatti et al. \citep{bonatti2007integration} and discuss two common approaches to trust collection and evaluation: policy-based and reputation-based trust management. Then, we propose our approach of Bayesian trust evaluation, which is a combination of policy-based and reputation-based methods. 

\paragraph{Policy-based Method:}
Policy-based methods enable the system to manage trust based on a set of predefined policies. These policies may include rules that specify the types of users or devices that are allowed to access certain resources, the level of access that is granted, and the conditions under which access is granted or denied. We provide several examples under this category.

\begin{itemize}
    \item \textbf{Network credential.}  The access request can be granted based on the given credentials of the entity. The trust information of the entity is encrypted in the credential as we assume only the trusted entity will process the credential. Kerberos \citep{neuman1994kerberos} is one example of authenticating service requests between trusted hosts across an untrusted network, such as the Internet. The underlying requirement for this method is that the system needs to ensure that the credential is private and not revealed to the attacker.
    
    \item \textbf{Ad-hoc attributes check.} The system can configure a set of qualified attributes that must be met before access is allowed. These attributes may include the device configuration, network environment security, application permission, etc. Identifying the necessary security attributes requires extensive knowledge of the system vulnerabilities. Poor security checks can result in inaccurate estimation of TS along with unresolved security vulnerabilities. 
    
    \item \textbf{Promise and incentive compliance.} Trust can be influenced by promises and penalties. To encourage desirable safe behaviors of the entity, the system can create a set of rules or contracts that promote incentive-compatible actions. A reward and penalty mechanism can also be strategically designed to elicit such behaviors. For instance, trust-based collaborative intrusion detection systems use incentive-compatible mechanisms to ensure no free-riding and facilitate cooperative network defense \cite{zhu2012guidex,fung2016facid}. Similarly, strategic trust frameworks based on evaluating the incentives of the opponents are used to guide the integration of IoT into communication networks \cite{pawlick2018istrict,pawlick2017strategic}.

\end{itemize}

\paragraph{Reputation-based Method:}

Reputation-based methods estimate the trustworthiness of an entity and adjust access permissions based on interactions or observations from past experiences, either directly (e.g., using historical behaviors) or indirectly (e.g., using third-party recommendation). This method can integrate more information but is also more vulnerable to false positives or false negatives. evaluation. We provide some examples under this category.

\begin{itemize}
    \item \textbf{Historical behaviors.} If the system has had direct interactions with the entity, the TS of the entity at this request can be developed based on the history of their encounters. The behavior characteristics of the entity can be multi-dimensional that involve login information, operational habits, abnormal behavior record, etc. The system needs to find proper risk measures that achieve the security goals. In addition, expired experiences should be excluded from the trust evaluation due to the dynamic features of modern networks. The out-of-date interaction record contributes little to the current trustworthiness of the entity and it is important to consider the attenuation in the data history.

    \item \textbf{Social Reputation.} Reputation from third parties or society can also serve as a source for trust evaluation.  Reputation may be defined as the global perception of the entity as being trustworthy. In other words, it is a collective trust opinion of other systems about the behavior of a subject node. The system prefers to grant access to a well-reputed entity. This information is helpful when the entity aims to enter the system for the first time.

    \item \textbf{Recommendation.} Recommendation is the simplest case of trust propagation. For instance, a recommendation from a trusted neighbor will increase the TS of the new entity. Reliable recommendation reduces information overload, uncertainties, and risk of the access attempt. It is important to provide a trust inference model to find a reliable recommendation that could improve trust evaluation accuracy.
    
    \item \textbf{Supply Chain.} Supply chain contains inter-organizational relationships among interdependent companies contributing to the final components in the target system. The trust in the suppliers also influence the trust evaluation of the device they provided. This type of trust propagation is multi-hop due to the multi-tier structure in the supply chain. Accountability investigation and cyber insurance in the supply chain \citep{ge2022accountability} could encourage truth-telling and information transparency to support trust evaluation.

    \item \textbf{Third-party Evaluation.} Besides the trust information propagated from others, the security system can also leverage third-party evaluation results (e.g., Intrusion Detection System (IDS) \citep{zhu2012guidex}, Security Information and Event Management (SIEM) \citep{miller2011security}, etc.) to develop a more reliable measure of trust score of the entity. The trace of the user provides a sequence of events that can be used for security analysis. It should be noted that the reliability of the side evidence  largely impacts trust propagation. The system needs to incorporate reliable side evidence for an accurate trust measure.
\end{itemize}

\paragraph{Bayesian Trust Evaluation}

Under the dynamic network environment, it is important for zero-trust security to continuously adjust the TS after the initial trust evaluation. The system needs to respond to changes in trust by investigating and orchestrating responses to potential incidents. The dynamic update should take account of previous knowledge about the entity as well as currently observed behaviors. In this chapter, we propose a Bayesian trust model to update the trust score. This model offers a quantitative way to combine policy-based trust with reputation-based evidence and update the TS subject to the perceived strategies of the entity.

\begin{definition}[\textbf{Bayesian Trust Update}]
The Trust Score (TS) of the entity $i$ at time $t+1$ is the probability that the entity is non-adversarial ($\theta^{t+1}_i\in\Theta_T$) based on the prior knowledge, side evidence, and observed strategies of the entity:
\begin{align}
    TS^{t+1}(i)=& \Pr(\theta_i^{t+1}\in\Theta_T|a^t,e^t,\pi^t) \notag \\
    =& \frac{h(e^t|a^t,\theta^t_i\in\Theta_T)\sigma(a^t|\theta^t_i\in\Theta_T)\pi^t(\theta^t_i\in\Theta_T)}{\sum_{\hat{\theta}_i\in\Theta} h(e^t|a^t,\hat{\theta}_i)\sigma(a^t|\hat{\theta}_i)\pi^t(\hat{\theta}_i)}
    \label{eq:bts}
\end{align}
where $a^t\in\mathcal{A}$ is the observed action of the entity, $e^t\in\mathcal{E}$ is the received side evidence, and $\pi^t$ is the system's prior knowledge about the entity up to time $t$. $\sigma$ is the observed strategy of the opponent and $h$ is the evidence-generating function given by the third party. Note that the relationship between $TS$ and $\pi$ is: $TS^t(i) = \pi^t(\theta^t_i\in\Theta_T)$.
\label{def:BTS}
\end{definition}

\begin{itemize}
    \item \textbf{Prior Knowledge $\pi^t$:} 
    Prior knowledge is the ex-ante likelihood of the entity being non-adversarial before taking into consideration any new (posterior) information. This information can be collected through various sources through policy-based methods or reputation-based methods. The initial trust score $TS^0(i)=\pi^0$ is usually constructed based on some kind of experience with, or firsthand knowledge of, the other party. For instance, attributes check, historical behaviors, reputation, etc. can all contribute to the prior computation. The system can establish an initial trust estimation of the entity at $t=0$ and compute the probability of the agent being trusted, i.e., $TS^0(i)\in [0,1]$. 
    
    \item \textbf{Side Evidence $e^t\in\mathcal{E}$:} The system can also incorporate side security evidence during trust updates. The side evidence may involve real-time network detection, system monitoring information, intelligent risk analysis, security alerts, etc. Reliable evidence helps establish a fast and accurate trust evaluation \citep{ge2022mufaza}.

    For instance, the external evidence is additional information taking binary value $e^t\in \mathcal{E}= \{0,1\}$, where $e^t=1$ indicates a security alarm, and $e^t=0$ means no alarm.  The security alarm warns the defender when the agent is more likely to be malicious. In general, the evidence is generated based on the probability that the agent with type $\theta^t$ takes an action $a^t$ at the current time $t$. Observing the evidence $e^t$, the defender can further update the trust of the agent via Bayes' rule.

    \item \textbf{Observed Strategies $\sigma(a^t|\theta^t)$: } The trustworthiness of an agent is determined by various factors, with their observed strategies playing a major part in trust updates. A strategy is a plan of actions that an agent intends to take to achieve its objectives. It also takes into account how an agent with a different type would behave in the current security state. Abnormal behaviors of the agent, such as attempting to access sensitive or restricted information in the system, could indicate that the agent has been compromised by an attacker. In such cases, the trust score of the agent should be decreased. To ensure the security of the system, it is essential to monitor the agent's behavior regularly and update or re-evaluate its trustworthiness as needed. 
\end{itemize}

\subsubsection{Purpose of Trust}
In this thesis, the TS is used to assist trust-based security policies. It plays a key role in establishing secure communication between different systems, networks, and individuals. Depending on the needs of the system, each situation places different requirements on trust. For instance, in data communication, the trust requirements would focus on the security level of the transmission environment.  In contrast, in supply chain security, the trust of the supplier depends more on the supplier's reputation and compliant behaviors.

\subsubsection{Trust Management}

Two major approaches to managing trust in cyber security are centralized and distributed trust management. Centralized trust management involves a central authority or entity that is responsible for managing and enforcing trust policies across a system or network \citep{ge2022trust,ge2023scenario}. It is typically used in environments where there is a clear hierarchy of trust relationships. Distributed trust management, on the other hand, involves a decentralized network of entities that are responsible for managing and enforcing trust policies. In this approach, trust decisions are made based on consensus among multiple entities, rather than by a single central authority. Each entity in the network may have its own trust policies and evaluation criteria, and trust decisions are made based on the collective evaluation of these policies and criteria. 

Both centralized and distributed trust management approaches have their advantages and disadvantages. Centralized trust management can provide a clear hierarchy of trust relationships and centralized enforcement of trust policies, but it may also be vulnerable to single points of failure and may require significant resources to maintain. Distributed trust management can be more resilient and adaptable to changing trust relationships, but it may also be more difficult to manage. The choice between centralized and distributed trust management in zero trust depends on the specific needs and requirements of the system or network.

\subsection{Game-Theoretic Trust}

Game theory plays a critical role in risk assessment by offering a structured framework to analyze and predict the outcomes of strategic interactions between attackers and defenders \cite{zhu2024foundations,rass2018game}. In cyber resilience, traditional risk assessment approaches often fall short because they focus on probabilistic measures without considering the intelligent and adaptive behaviors of adversaries. Game theory fills this gap by introducing the  of strategic cyber risk, where risks are not merely quantified by the likelihood of certain events but also by the actions, goals, and adaptive strategies of both attackers and defenders. By modeling these interactions as a game, defenders can better predict likely attack strategies, optimize their defense mechanisms, and allocate resources more effectively.

In this context, game theory allows defenders to model risk using dynamic interactions \cite{zhang2020game,huang2020dynamic,huang2019dynamic}. Cyber threats evolve over time, making static risk assessments insufficient. Game theory addresses this issue by using dynamic game models that capture the ongoing interaction between attackers and defenders. For instance, repeated games model persistent threats where adversaries continuously probe a system for weaknesses, while sequential games capture how both attackers and defenders adjust their strategies over multiple stages of interaction. In these models, resilience mechanisms can be deployed in three stages: proactive measures to prevent attacks, responsive mechanisms to react in real-time, and retrospective strategies to recover from damage already done. Each stage is modeled in terms of payoffs (the consequences of actions) and transition dynamics (how actions change the system’s state). This dynamic perspective enables defenders to adapt their strategies based on observed behaviors and real-time threats, enhancing overall cyber resilience.

A key feature of game theory in risk assessment is its ability to handle situations with asymmetric information \cite{li2023price,chen2018linear,li2024conjectural,li2022commitment}. Often, attackers have more information about certain vulnerabilities, or defenders may not fully understand the attacker’s capabilities. In such cases, game theory uses models like Bayesian games and signaling games. Bayesian games are particularly useful in scenarios where players have incomplete information about each other’s actions and intentions. For example, a defender may not know the exact nature of an attacker’s capabilities or targets, but they can infer these based on the prior information.  More formally let's consider that each agent \( i \) has a private type \( \theta_i \), which reflects characteristics relevant to trustworthiness (e.g., history of reliable behavior). The set of all possible types for agent \( i \) is \( \Theta_i \). The joint distribution over types for all agents is denoted by \( p(\theta) \), where \( \theta = (\theta_1, \theta_2, \ldots, \theta_N) \) represents the type profile of all agents. Since each agent is unaware of the exact types of others, they rely on their beliefs about the type distribution, denoted by \( p(\theta_{-i}|\theta_i) \), where \( \theta_{-i} \) represents the types of agents other than \( i \). In a trust evaluation framework, agent \( i \) forms a strategy based on their beliefs about others' types, aiming to maximize their expected utility given the probabilistic nature of their beliefs. Agent \( i \)'s utility function \( u_i(a, \theta) \) depends on both the action profile \( a = (a_1, a_2, \ldots, a_N) \) and the type profile \( \theta \). Agent \( i \)’s expected utility, given their type \( \theta_i \) and belief \( p(\theta_{-i}|\theta_i) \), is calculated as:

\[
\mathbb{E}[u_i(a, \theta) | \theta_i] = \sum_{\theta_{-i} \in \Theta_{-i}} p(\theta_{-i}|\theta_i) u_i(a, \theta).
\]
This expectation integrates over all possible types of other agents, weighted by agent \( i \)’s belief in each type, providing a trust-informed utility estimate.

In game-theoretic trust settings, Bayesian Nash Equilibrium (BNE) represents a stable outcome where each agent \( i \) chooses a strategy \( \sigma_i^* \) that maximizes their expected utility given their type and their beliefs about other agents' strategies and types. Formally, a BNE for agent \( i \) with type \( \theta_i \) is achieved when:

\[
\sigma_i^*(\theta_i) \in \arg \max_{a_i} \sum_{\theta_{-i} \in \Theta_{-i}} p(\theta_{-i}|\theta_i) u_i\left(a_i, \sigma_{-i}^*(\theta_{-i}), \theta\right).
\]

Recent research has explored the use of Bayesian-based beliefs and strategies to differentiate legitimate users from potential attackers. The GAZETA framework \cite{ge2023gazeta} proposes a game-theoretic, zero-trust authentication schema that employs dynamic game models to enable zero-trust defense in networked systems. This framework supports a robust, resilient, and efficient zero-trust model, leveraging game theory and trust for enhanced cybersecurity.

Signaling games are a class of dynamic Bayesian games. They are essential for trust evaluation scenarios where agents must decide whether to trust others based on observed signals and prior beliefs. In these games, agents act as either senders or receivers of signals, and their strategies involve both sending and interpreting signals to maximize expected utility under conditions of incomplete information. In a signaling game for trust evaluation, a sender with type \( \theta_i \) chooses a signal \( s \in S \) to send to a receiver. The receiver, who does not know the sender's type, updates their belief about \( \theta_i \) using Bayes’ rule based on the observed signal \( s \). If \( p(\theta_i) \) is the prior probability of the sender’s type, then the receiver’s posterior belief after observing \( s \) is:

\[
p(\theta_i | s) = \frac{p(s | \theta_i) p(\theta_i)}{\sum_{\theta_i' \in \Theta_i} p(s | \theta_i') p(\theta_i')}.
\]

This updated belief \( p(\theta_i | s) \) reflects the receiver’s revised confidence in the sender’s intentions or trustworthiness based on the signal received. Once the receiver updates their belief about the sender’s type, they must decide whether to trust the sender by taking an action \( a_j \) (where \( j \) denotes the receiver) that maximizes their expected utility. The receiver’s expected utility, conditional on the updated belief \( p(\theta_i | s) \), is:

\[
\mathbb{E}[u_j(a_j, s) | \theta_j] = \sum_{\theta_i} p(\theta_i | s) u_j(a_j, \theta_i),
\]
where \( u_j(a_j, \theta_i) \) is the utility the receiver gains from taking action \( a_j \) given their belief about the sender’s type \( \theta_i \). This framework allows the receiver to weigh the benefits of trusting the sender against the potential risks.

In signaling games, the sender anticipates how the receiver will interpret signals and selects a signal \( s \) to maximize their own expected utility. The sender’s strategy, \( \sigma_i^*(\theta_i) \), is chosen to optimize their outcome by influencing the receiver’s trust decision. The sender’s optimal strategy is:

\[
\sigma_i^*(\theta_i) = \arg \max_{s} \sum_{a_j} p(a_j | s) u_i(s, a_j, \theta_i),
\]
where \( u_i(s, a_j, \theta_i) \) is the utility for the sender given their type \( \theta_i \), the chosen signal \( s \), and the receiver’s action \( a_j \). The equilibrium in signaling games is often referred to as a Perfect Bayesian Equilibrium (PBE), in which (i) The sender’s signal \( s \) is optimally chosen based on their type \( \theta_i \); (ii) The receiver’s action \( a_j \) is optimal given the posterior belief \( p(\theta_i | s) \) and their expected utility.

In the context of cyber deception, signaling games can be applied to scenarios where defenders (senders) send misleading signals to manipulate attackers’ (receivers) beliefs and actions, steering them toward suboptimal choices. For example, a defender may deploy a honeypot—a deceptive signal intended to appear as a vulnerable system. The attacker observes this signal \( s = \text{“honeypot”} \) and updates their belief about the system’s risk level \cite{pawlick2015deception}. If the attacker perceives the system as low-risk, they may proceed with an attack, revealing themselves and wasting resources. %The defender’s expected utility \( u_j(a_j, s) \) from deploying a honeypot can be represented as:
%
%\[
%\mathbb{E}[u_j(a_j, s) | \theta_j] = \sum_{\theta_i} p(\theta_i | s) u_j(a_j, \theta_i),
%\]
%where \( \theta_i \) is the attacker’s type, \( a_j \) is the defender’s action, and \( u_j(a_j, \theta_i) \) is the utility derived from influencing the attacker’s choice.
%
%In a cyber deception signaling game, the attacker (now acting as the receiver) must decide on an action \( a_i \) to maximize their own utility, taking into account the possibility that the signal \( s \) may be deceptive. The attacker’s expected utility, given the observed signal and belief about the defender’s intentions, can be represented as:
%
%\[
%\mathbb{E}[u_i(a_i, s) | \theta_i] = \sum_{\theta_j} p(\theta_j | s) u_i(a_i, \theta_j),
%\]
%where \( \theta_j \) is the defender’s type, influencing the attacker’s perception of the signal’s reliability.
% 
In a Perfect Bayesian Equilibrium (PBE), both the sender’s and receiver’s strategies are optimal and consistent with their beliefs and the observed actions of the other party. This equilibrium framework in signaling games enables robust trust evaluation and effective cyber deception by aligning agents’ actions with their updated beliefs, allowing them to balance potential rewards and risks effectively.

Game theory enhances trust and risk assessment in cybersecurity through payoff functions that quantify the costs and benefits of actions for both attackers and defenders. For defenders, payoffs account for the costs of implementing security measures, such as network segmentation or cyber deception, and the potential financial or reputational damage from a successful attack. For attackers, the payoff represents the value they gain from exploiting system vulnerabilities. The overall cyber risk can be dynamically assessed by calculating the expected payoff, which is often modeled as the product of the attack’s success probability and the impact magnitude. This dynamic risk profile evolves as both attackers and defenders adjust their strategies, capturing the adaptive nature of cyber conflict rather than relying on static risk probabilities.

Another essential contribution of game theory to trust and risk assessment is mechanism design \cite{zhang2021informational}, where defenders proactively shape the rules of engagement to influence attacker behavior. Mechanism design frequently utilizes Stackelberg games \cite{liu2024stackelberg}, where the defender (acting as the leader) anticipates the attacker’s (follower’s) responses and designs the system to channel the attacker towards less harmful actions. An example is cyber insurance \cite{liu2022role}, where game theory aids in evaluating network compromise risks and designing policies that incentivize organizations to adopt stronger security measures. Similarly, in zero-trust architectures \cite{ge2023gazeta}, game theory supports adaptive access control by continuously monitoring user behavior and adjusting authentication requirements, thereby minimizing the attacker’s advantage and enhancing system resilience.

Game theory can also incorporate bounded rationality and learning algorithms \cite{rass2020bounded}, recognizing that attackers and defenders often operate with limited information and computational resources. In practical settings, adversaries may not act with full rationality, and defenders may lack complete models of attacker behavior. Game theory addresses these limitations with models of bounded rationality and reinforcement learning \cite{li2022role}, allowing both sides to learn and adjust their strategies over time. For example, conjectural learning \cite{li2024conjectural} enables defenders to form hypotheses about future attacker actions based on observed past behavior, allowing continuous refinement of defense strategies. These adaptive learning models create a robust framework for trust and risk assessment, accommodating the evolving strategies and imperfect information typical in real-world cybersecurity scenarios.

\section{Role of Game Theory in Strengthening AI Trustworthiness}
\label{sec:strengthen}
The main issues with AI security revolve around the growing vulnerabilities created by the integration of AI into a wide range of systems, significantly increasing the attack surface. Traditional cyber defenses are often not equipped to handle these new attack vectors, particularly in machine learning (ML)-based AI systems, which are highly susceptible to adversarial attacks. In these attacks, inputs are intentionally manipulated to deceive the AI models, causing them to make incorrect predictions or decisions \cite{ge2023ai}. These manipulations can target various AI-enabled systems, including those used in facial recognition, healthcare, and autonomous vehicles. Real-world case studies have shown that adversarial attacks can result in severe financial damage, such as a notable case where a facial recognition system suffered millions in losses due to an adversarial attack.

One of the most pressing challenges in AI security is bias. AI models can inadvertently learn biases from the datasets they are trained on, leading to unsafe, unfair, or discriminatory outcomes \cite{ge2024attributing}. Bias is not only a social concern but also a security risk because it can be exploited by adversaries to degrade the performance of the system or manipulate its behavior. Ensuring the trustworthiness of AI models is crucial for making them reliable in high-stakes scenarios, particularly where fairness, accountability, and transparency are critical.

A major concern exacerbating AI security issues is the increased attack surface. The integration of AI into existing infrastructure expands the number of possible entry points for attackers. AI systems interact with vast data sets and complex networks, making them more vulnerable to both traditional cyberattacks and AI-specific exploits. These expanded attack surfaces include the data pipelines feeding into AI models, the models themselves, and the environments in which these models operate. This introduces new vectors for attacks, such as data poisoning and model inversion, which can compromise the integrity and confidentiality of AI systems.

\subsection{Robust Adversarial Training}
One of the most effective ways to counter adversarial attacks is through adversarial training. This approach involves training AI models on adversarial examples—inputs that have been intentionally manipulated to test the model’s robustness. By exposing models to these adversarial examples during the training phase, the AI can learn to recognize and resist such manipulations in real-world scenarios. Adversarial training helps build a model’s resilience to deceptive inputs, improving its performance in the face of malicious attacks.

Game theory plays a foundational role in adversarial training by modeling the ongoing strategic conflict between adversaries (attackers) and machine learning models (defenders). This framework facilitates understanding how machine learning systems can be made more resilient to adversarial attacks, particularly those that seek to exploit weaknesses in AI models by introducing deceptive inputs. Game theory allows researchers to model this as a dynamic, multi-step interaction between two competing agents, leading to better defense strategies and increased robustness of the models.

\subsubsection{Adversarial Training as a Min-Max Game}
In adversarial training, the goal is to harden models against adversarial perturbations—small, imperceptible changes to input data that can lead models to make incorrect predictions. This process is typically modeled as a min-max optimization problem \cite{razaviyayn2020nonconvex,ren2021towards}, where the adversary seeks to maximize the model's classification error (or loss), and the defender (the model) seeks to minimize this error under worst-case perturbations. Specifically, the adversary creates adversarial examples to fool the model, while the defender tries to adapt by learning from these examples and improving the model's resilience. This iterative optimization can be described as a two-player zero-sum game, where the adversary’s gain is directly proportional to the model’s loss.

Game theory helps  formalize this relationship and structure the learning process. By modeling the training as a game, it becomes possible to assess the effectiveness of the model in withstanding worst-case adversarial attacks. The game-theoretic approach allows for the exploration of equilibrium points—where neither the attacker nor the defender can further improve their position without changing their strategy—helping to identify optimal defense mechanisms.

Furthermore, game theory allows for the analysis of more sophisticated attack-defense dynamics, such as scenarios involving multiple adversaries or defenders, each with their own objectives and constraints. This multi-agent framework can be extended to incorporate distributional adversarial training, where the defender learns to counter a broad distribution of potential attacks rather than focusing on a specific type of adversarial example \cite{kim2020adversarial, ye2021towards}.
 
Recent research has sought to provide a unified game-theoretic interpretation of adversarial perturbations and robustness, helping to explain the behavior of adversarially trained models and offering insights into why certain defense strategies succeed. This framework suggests that adversarial perturbations primarily affect high-order interactions in deep neural networks, while adversarial training helps models build resilience by focusing on low-order interactions that are more robust to attack \cite{ye2021towards}.

\subsubsection{Stackelberg Games in Adversarial Learning}
A common game-theoretic framework used in adversarial training is the Stackelberg game \cite{bruckner2011stackelberg,li2024meta,liu2024stackelberg,pawlick2016stackelberg}. In this framework, the adversary is modeled as the "leader" who makes the first move by crafting adversarial examples, and the defender is the "follower" who responds by updating the model to minimize the impact of these examples. The Stackelberg model is particularly useful because it accounts for the sequential nature of attacks and defenses, capturing the dynamic, iterative nature of adversarial learning.

In a Stackelberg game, the leader (the adversary) optimizes their strategy with full knowledge that the follower (the defender) will react to their move. The defender then optimizes their strategy based on the adversary’s action. This setup provides a way to formally analyze the adversary’s behavior and design more effective defense strategies. The Nash equilibrium of the Stackelberg game represents a point where neither the adversary nor the defender can improve their outcome by changing their strategies, making it a stable state for the defense process.

\subsubsection{Dynamic Interactions in Adversarial Training}
Adversarial training often involves alternating between generating adversarial examples and updating the model \cite{zhang2016dual,zhang2016dynamic,zhang2021security}. In game-theoretic terms, this is called an alternating best-response strategy. The adversary first generates an example that maximizes the model’s loss, and then the model updates its parameters to minimize this loss. This process continues in a loop until an equilibrium is reached. The defender's updates correspond to minimizing the loss over a worst-case adversarial distribution, which can be viewed as solving a min-max game at each iteration.

However, challenges arise because this adversarial process may not always converge smoothly. As noted in some studies, alternating best-response strategies can lead to non-converging behavior, especially when the game is not convex-concave \cite{razaviyayn2020nonconvex}. This non-convergence can make it difficult to find robust solutions, as the iterative game between the attacker and the defender might cycle indefinitely without reaching an equilibrium. Game theory helps in understanding these dynamics and suggesting conditions under which convergence can be achieved, as well as identifying Nash equilibria that guarantee robust solutions .

\subsection{Red and Blue Teaming}
Another crucial domain in AI security is red teaming and threat emulation.  AI red teaming simulates adversarial behavior to rigorously test system defenses, identifying vulnerabilities before real-world attackers can exploit them. Frameworks like MITRE’s ATLAS equip AI developers with insights to anticipate potential threats and refine defenses. Through threat emulation techniques, security teams mimic real-world attack scenarios to evaluate an AI system's resilience against complex, evolving threats.

Game theory forms the backbone of red and blue teaming dynamics, where red teams adopt an offensive role as attackers and blue teams defend the system. Integrating these approaches into a purple teaming strategy enables organizations to simulate a comprehensive, cyclical attack-defense process, bolstering system resilience by operationalizing adversarial insights. Innovations like PenHeal \cite{huang2024penheal} showcase how game theory and agent-based LLM solutions can strategically design and execute security defenses through simulated adversarial and defensive interactions. Combined with foundation models—large, pre-trained AI models—these solutions operationalize complex threat detection and mitigation strategies in real-time.

The recently proposed ADAPT framework \cite{huang2019adaptive} provides a powerful example of game-theoretic principles in action. ADAPT is designed to tackle vulnerabilities in AI-driven systems, especially within complex, high-stakes infrastructures like healthcare. Leveraging meta- and micro-games, ADAPT facilitates automated penetration testing that simulates adversarial scenarios at different levels, making it foundational to advanced AI red teaming and enhancing the security posture of AI systems against real-world threats.

\subsubsection{Case Study: Securing AI-Driven Traffic Management System Using the MITRE ATLAS Framework}

A large metropolitan city deployed an AI-driven traffic management system to optimize traffic flow, reduce congestion, and enhance public safety. This system integrates machine learning algorithms, sensor data from vehicles and cameras, and GPS information to make real-time adjustments to traffic signals and manage road congestion. However, given the critical role of this system in the city’s infrastructure, it became a prime target for adversarial cyberattacks aimed at disrupting traffic patterns and creating chaos, particularly during emergency situations.

During a routine red teaming exercise, a simulated cyberattack was executed, targeting the traffic management AI system. The goal was to manipulate the system’s decision-making process through data poisoning. The red team fed the AI system corrupted sensor and GPS data, leading the AI to misclassify traffic conditions. This misinterpretation resulted in improper traffic signal adjustments, causing gridlock at major intersections, delays in emergency response times, and widespread traffic disruption throughout the city.

Moreover, the attackers employed model evasion techniques, where subtle modifications to input data allowed them to bypass the AI system’s security mechanisms. These adversarial perturbations were designed to be undetectable but effective enough to influence the system’s decisions. This type of attack was based on real-world adversarial tactics cataloged by the MITRE ATLAS framework \cite{mitreatlas}, which documents vulnerabilities specific to AI systems in critical infrastructures.

\subsubsection{Attack Techniques:}

The red team utilized several advanced techniques outlined in the MITRE ATLAS framework, including:

\begin{itemize}
\item Data poisoning is one of the most impactful attacks on AI systems, where adversaries inject malicious data into the training set, leading the AI to learn incorrect patterns. This can cause severe misclassifications in real-time operations. The red team in this case study introduced corrupted GPS and sensor data, causing the AI to misinterpret traffic conditions, ultimately resulting in traffic mismanagement. For example, the Common Vulnerabilities and Exposures (CVE) system includes vulnerabilities like CVE-2021-28370, which highlights a weakness where poisoned datasets could cause AI models to misbehave. The MITRE ATLAS framework helps in understanding how to detect and mitigate these data poisoning attacks by ensuring that training data is properly verified and tested before being applied in real-world AI models.

\item Adversarial perturbations involve introducing small modifications to input data, which cause AI systems to make erroneous decisions. In the traffic management case study, the red team slightly altered the sensor and GPS inputs in such a way that the system incorrectly interpreted traffic patterns. The perturbations were designed to evade basic security checks but still manipulate the system’s behavior. MITRE ATLAS emphasizes this technique by categorizing such attacks under adversarial examples that deceive models into making false predictions. For instance, CVE-2020-14472 describes how adversarial examples can be created to evade detection, which is closely aligned with this attack method. Mitigation strategies include defensive distillation, which makes the model less sensitive to such minor modifications.

\item  Model evasion refers to attacks that allow adversaries to bypass the AI system’s security defenses without detection. The red team in the traffic management scenario used this technique to adjust inputs in a way that avoided triggering alarms while still influencing the AI’s decision-making process. The adversaries employed evasion strategies to manipulate the traffic light timing and create gridlocks. Model evasion attacks are well-documented in MITRE ATLAS and are also referenced in specific CVE entries, such as CVE-2020-10713, which outlines vulnerabilities where evasion techniques can bypass AI security checks. Mitigation involves improving anomaly detection systems and employing stronger model verification processes.

 \end{itemize}
 
These adversarial techniques are precisely mapped in the MITRE ATLAS framework, which categorizes attack vectors like data poisoning, adversarial examples, and model evasion under a structured set of Tactics, Techniques, and Procedures (TTPs). MITRE ATLAS serves as a critical tool for AI developers and security professionals to better understand and anticipate adversarial behaviors, helping to enhance the defense posture of AI systems. By utilizing such frameworks, AI-driven systems, like the one in the   case study, can anticipate and defend against sophisticated adversarial threats.

In this case study, red teaming and threat emulation played a critical role in testing and securing the AI-driven traffic management system. Red teaming involves simulating adversarial behaviors to identify vulnerabilities before real-world attackers can exploit them. By using tools from the MITRE ATLAS framework, the red team anticipated potential threats and developed attack strategies such as data poisoning and model evasion. These techniques allowed the team to feed corrupted data into the system, causing disruptions in traffic management decisions, such as improper signal timing and gridlocks.

The threat emulation approach, which mimics real-world attack scenarios, enabled the red team to explore how small, seemingly innocuous perturbations in data could bypass security mechanisms, leading to widespread system failure. This process provided valuable insights for the blue team (defenders), who responded by applying countermeasures such as adversarial training and defensive distillation. These AI-specific defenses hardened the system, making it more resistant to adversarial manipulations in the future. The use of game theory is fundamental in underpinning the adversarial dynamics between red and blue teams. Game theory models the interaction as a strategic game where both teams adapt their strategies in response to the actions of the other. The red team (attackers) continuously seeks to maximize disruption, while the blue team (defenders) aims to minimize damage and maintain system integrity. Through game-theoretic modeling and simulations, both teams were able to anticipate each other's moves and optimize their strategies.

In this case, we can adopt a purple teaming approach, where red and blue teams collaborate  to simulate a comprehensive attack-defense cycle. Purple teaming enables the organization to test and refine both offensive and defensive strategies iteratively, creating a feedback loop that continuously improves system resilience. By simulating realistic attack scenarios and assessing defenses in real-time, the organization can operationalize these strategies more effectively. Further reinforcing this approach,  game-theoretic models can be used to strategically plan security defenses,  combining agent-based solutions with game theory to simulate both adversarial and defensive scenarios. These simulations allowed for more precise planning of responses to potential attacks, ensuring that both red and blue teams could adjust their strategies dynamically during the exercise.

Finally, the integration of foundation models, large pre-trained AI models, operationalized the system’s real-time defense. These models enhanced the system’s capability to detect and respond to adversarial threats as they evolved. The foundation models analyzed incoming traffic and sensor data, detected unusual patterns indicative of attacks, and automatically adjusted traffic signals and routes to mitigate the impact of the adversarial inputs. This combination of game theory, red and blue teaming, and the use of foundation models ensured a resilient AI system capable of defending against increasingly sophisticated threats.

\section{Conclusion}
\label{sec:conclusion}

In conclusion, the chapter underscores the intertwined nature of AI and trust in achieving secure, resilient networked systems. AI enables advanced trust management through real-time adaptability and strategic insight, yet its deployment requires a strong foundation of trust in the technology itself. Game theory emerges as a vital tool, modeling the adversarial dynamics and guiding adaptive trust mechanisms that allow AI to respond effectively to evolving cybersecurity threats. While AI advances trust evaluation, a governance framework that addresses transparency, accountability, and ethical use is essential to foster a positive feedback loop. By reinforcing trust in AI and leveraging AI to enhance network security, organizations can achieve a sustainable equilibrium that supports the long-term adoption and resilience of AI-powered systems.

\bibliographystyle{plainnat}
\bibliography{ref}

\end{document}